\documentclass[letterpaper]{article}
\usepackage{aaaicut}
\usepackage{times}
\usepackage{helvet}
\usepackage{courier}
\usepackage{algorithm,algorithmic}
\usepackage{xcolor}
\definecolor{myOrange}{rgb}{1,0.5,0}
\usepackage{times,epsfig,graphicx}
\usepackage{float}
\usepackage{amssymb,amsmath}
\usepackage[latin1]{inputenc}
\usepackage{fancyhdr}
\usepackage{verbatim}
\usepackage{tabularx}
\usepackage{hyperref}
\usepackage{bbm}
\usepackage{dsfont}
\usepackage{xspace}
\usepackage{bm}

\newlength{\smallimage}
        \definecolor{rel}{rgb}{.1,.6,.2}
        \definecolor{nrl}{rgb}{1,1,1}
        \definecolor{qim}{rgb}{1,1,1}

\makeatletter

\DeclareRobustCommand\onedot{\futurelet\@let@token\@onedot}
\def\@onedot{\ifx\@let@token.\else.\null\fi\xspace}

\newcommand{\rar}{\rightarrow}

\def\be{\begin{equation}}
\def\ee{\end{equation}}
\def\bea{\begin{eqnarray}}
\def\eea{\end{eqnarray}}
\def\ben{\begin{eqnarray*}}
\def\een{\end{eqnarray*}}

\def\bi{\begin{itemize}}
\def\ei{\end{itemize}}

\newcommand{\bt}[1]{\begin{tabular}{#1}}
\newcommand{\et}{\end{tabular}}
\newcommand{\ba}[1]{\begin{array}{#1}}
\newcommand{\ea}{\end{array}}

\def\<{\langle}
\def\>{\rangle}

\DeclareMathOperator*{\E}{\mathbb{E}}

\begin{document}
\title{Using Latent Codes for Class Imbalance Problem in Unsupervised Domain Adaptation}
\author{ 
Boris Chidlovskii \\
Naver Labs Europe, Meylan 38240, France}

\maketitle 
\begin{abstract}
We address the problem of severe class imbalance in unsupervised domain adaptation, when the class spaces in source and target domains diverge considerably. Till recently, domain adaptation methods assumed the aligned class spaces, such that reducing distribution divergence makes the transfer between domains easier. Such an alignment assumption is invalidated in real world scenarios where some source classes are often under-represented or simply absent in the target domain. We revise the current approaches to class imbalance and propose a new one that uses latent codes in the adversarial domain adaptation framework. We show how the latent codes can be used to disentangle the silent structure of the target domain and to identify under-represented classes. We show how to learn the latent code reconstruction jointly with the domain invariant representation and use them to accurately estimate the target labels.

\end{abstract}

\section{Introduction}
\label{sec:introduction}

Significant advances that deep neural networks achieved in various applications 
are to a large extend due to the availability of large-scale labeled data. As the manual annotation of massive 
training data remain expensive, 
domain adaptation strategies~\cite{wang18survey} 
have been proposed to leverage labeled data from different but related domains. These methods cope with the shift in data distributions across domains preventing predictive models from a generalization to new target tasks.



Unsupervised domain adaptation (UDA) aims at compensating the data shift between different domains by learning domain-invariant feature representations, using labeled source and unlabeled target domain data.

Most approaches rely on the comparison of marginal distributions between the source and target domains, 
and, till recently, they implicitly assumed a good class alignment 
between the source and target domains. 
Under such assumption, various families of domain adaptation methods have been developed; most popular families perform the adaptation by matching the statistic moments~\cite{long2017cond,long2016}, by optimal transportation between the domains~\cite{courty2017} or by relying on the domain adversarial nets~\cite{chadha2018,GaninJMLR16Domainadversarial,tzeng2017adda}. 

More practical scenario has been addressed by the open set case~\cite{Saito_2018_ECCV} and partial domain adaptation, where the target domain contains a subset of source classes~\cite{cao2018-cvpr,cao2018-eccv,chen_re-weighted2018,cao2019learning} 
%
In this paper, we make a step forward and study the domain adaptation under the {\it severe class imbalance}, which generalizes the absence of some classes in the target domain
to the continuous space of divergences between source and target class distributions.
A very relevant problem of {\it selective bias} has been studied in semi-supervised~\cite{zadrozny_learning_2004} and deep learning~\cite{ren_learning_2018}, so we 
re-use the evaluation protocol they developed to test the unsupervised domain adaptation methods under the class imbalance.

We measure the class imbalance ($CI$) as the Kullback-Leibler (KL) divergence of target class distribution $p_t(y)$ from the source class distribution $p_s(y)$. 
The KL divergence of 0 indicates a similar behavior of the two distributions, while a high KL divergence indicates that the distributions behave in very different manners. 

\begin{figure}[ht]
\centering{
\includegraphics[width=0.45\textwidth]{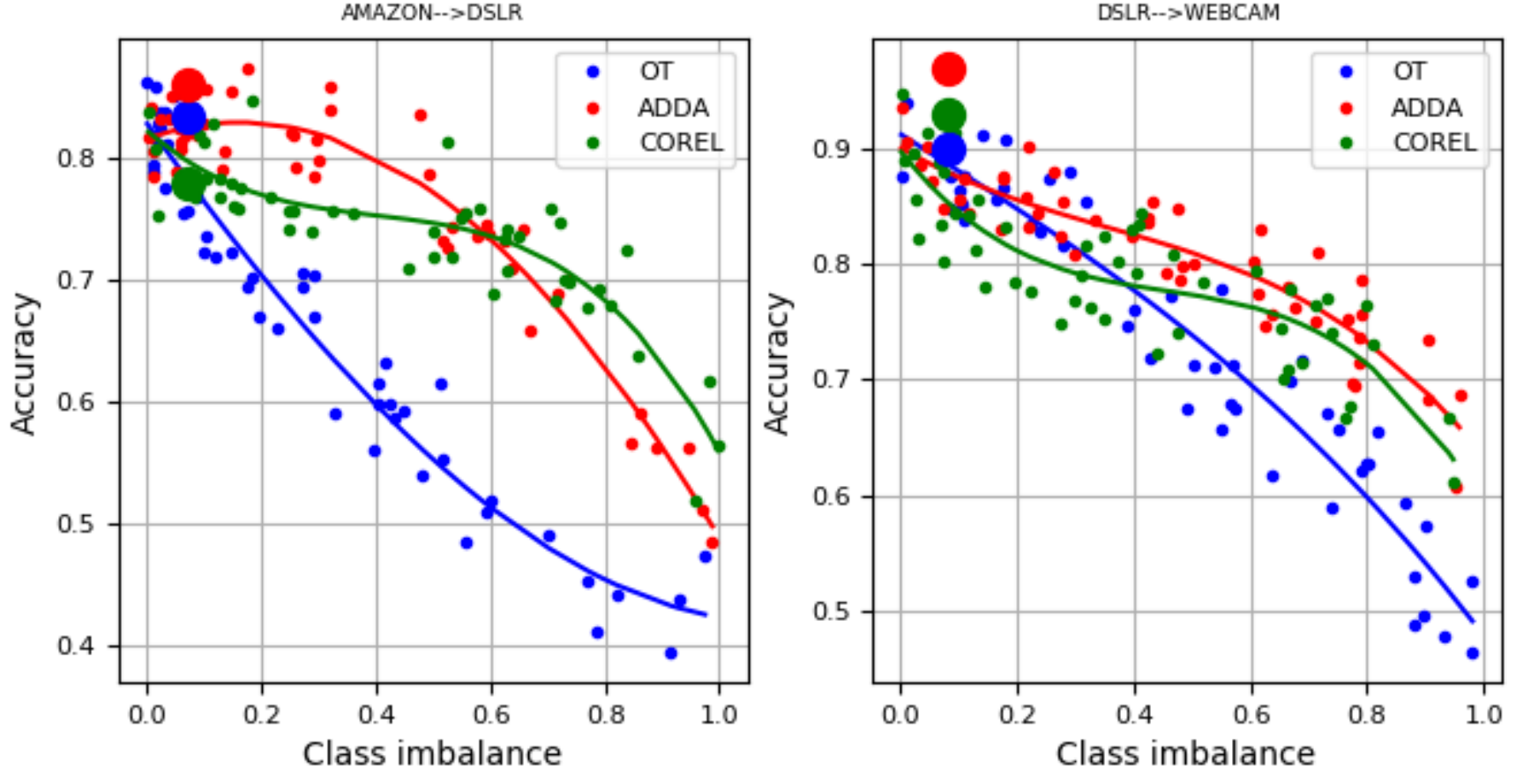}
}
\caption{Impact of class imbalance on accuracy of domain adaptation methods (better seen in color).}
\label{fig:ci}
\end{figure}

We 
test the resistance of three state of art domain adaptation methods, based on the optimal transport (OT)~\cite{courty2017}, adversarial discriminative domain adaptation (ADDA)~\cite{tzeng_adversarial_2017} and the correlation alignment (COREL)~\cite{sun2017}. Code of each method
has been modified to allow sampling the target domain with uniformly distributed $CI$ values, $CI\sim Unif(0,1)$. 
Figure~\ref{fig:ci} plots the classification accuracy in the function of $CI$ values, for three domain adaptation tasks from the standard Office31 dataset.
Large colored circles indicate the accuracy in the standard case of class alignment, where $CI$ values vary between 0.03 and 0.08. It is obvious that comparing the UDA methods for small $CI$  
says little about their resistance to a severe class imbalance. 
Meanwhile, when $CI$ values approach 1, the accuracy drops by 30\% to 50\%.

\begin{figure}[ht]
\centering{\includegraphics[width=0.5\textwidth]{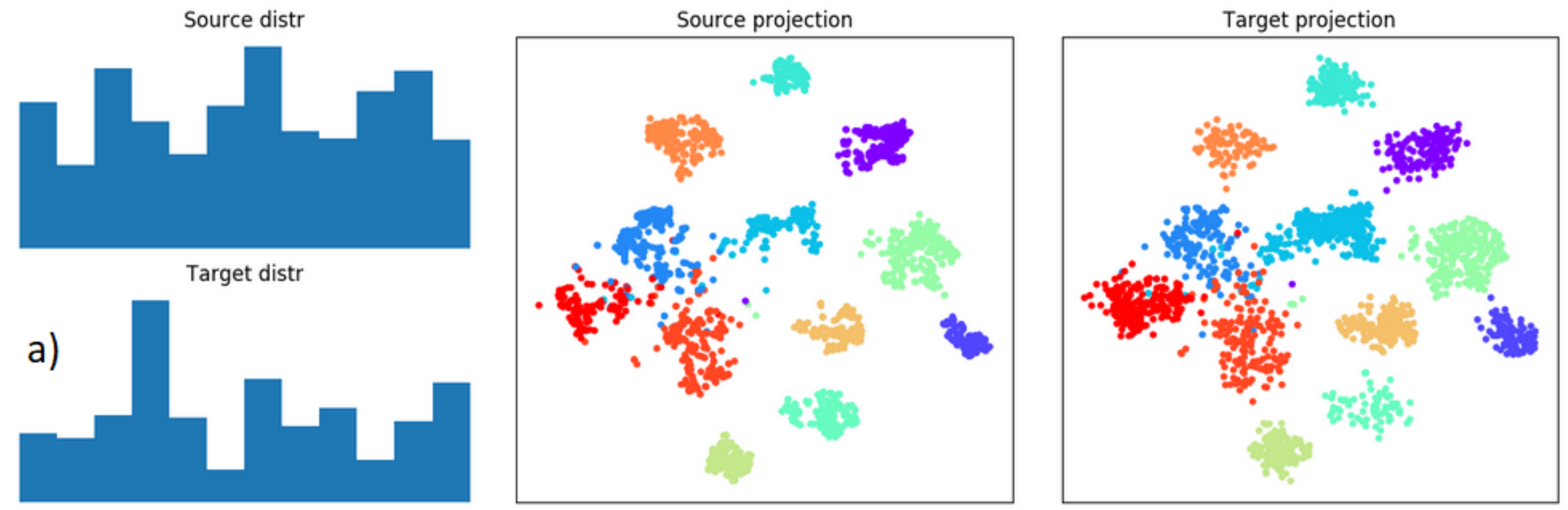}\\
\includegraphics[width=0.5\textwidth]{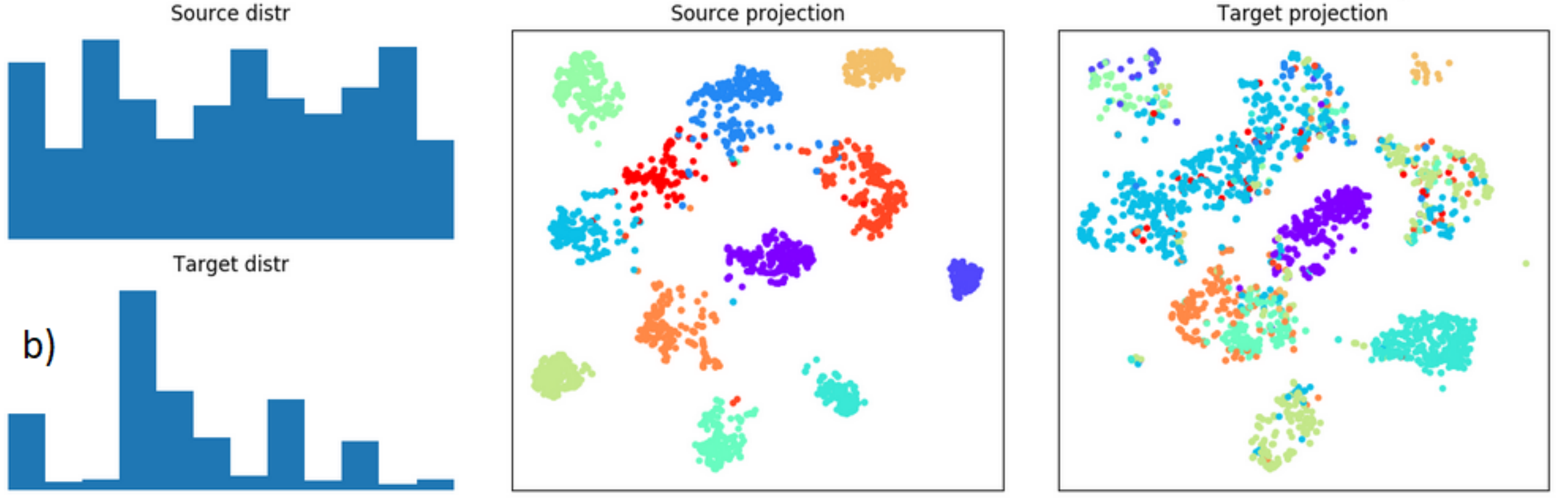}}
\caption{T-SNE projections of source and target samples, under (a) small ($CI$=0.11) and (b) significant ($CI$=0.71) class imbalance (better seen in color).} 
\label{fig:ci2}
\end{figure}

This performance drop can be explained using the t-SNE projections of source and target instances. When class imbalance is small (Figure~\ref{fig:ci2}.a, $CI$=0.11), the UDA methods easily separate the target classes. Instead, when the class imbalance becomes significant (Figure~\ref{fig:ci2}.b, $CI$=0.71), the methods fail to separate small and big classes.
Under severe class imbalance, it is impossible to reduce the domain shift by comparing source and target distributions directly. Reducing distribution shift will not benefit to the target task, since the marginal distributions between domains should not be the same due to the different label spaces. 

A natural way to transfer between domains is re-weighting the source domain samples whose classes are likely to appear in the target domain. 
As the target domain is unlabelled, it appears however challenging to uncover which classes are presented and which source domain samples are important for transferring. 
Success of any re-weighting schema relies on capacity to correctly estimate the target labels. Straightforward under the assumption of the class alignment, it becomes problematic in the case of severe class imbalance. 

There are two major approaches to cope with the class imbalance in the domain adaptation. One is based on the capacity of source classifier~\cite{cao2018-eccv,yan17} to make accurate target predictions, another uses the domain discriminator~\cite{hung2018} to detect which classes appear in the target domain.
We propose an alternative approach inspired by the conditional GANs. We inject  {\it the latent codes} in the adversarial discriminative 
framework, with the goal to disentangle the salient structure of the target domain.
We follow InfoGAN~\cite{chen_infogan:_2016} in how these latent codes can be reconstructed jointly with the learning the domain-invariant representations, and assist the source classifier to accurately estimate target class distribution. 


The InfoGAN loss includes a non-differentiable entropy term and its training mechanism assumes that the latent codes are sampled from a fixed distribution. Obvious for the source domain where the class distribution is known, such an assumption is invalidated in the target domain where the class distribution is unknown. To make the entropy term differentiable, we propose {\it a continuous relaxation} to the loss objective; it allows to integrate the estimation of target class distribution in the training process.

The remainder of this paper is organized as follows. Section~\ref{sec:soa} presents state of art methods on the unsupervised domain adaptation and the reviews the recent efforts to cope with the class imbalance. Section~\ref{sec:lada} describes the adversarial domain adaptation network and proposes a novel extension which uses the latent codes for disentangling the target domain and estimate the target class distribution. Section~\ref{sec:eval} presents Office31 and VisDA datasets and an extended evaluation protocol for testing UDA methods under the class imbalance; it then reports the evaluation results. Section~\ref{sec:conclusion} concludes the paper.
\section{State of Art}
\label{sec:soa}
The development of deep neural networks 
has boosted a large number of machine learning problems and 
computer vision applications. 
Deep neural networks can learn more transferable features, by disentangling explanatory factors of variations underlying data samples~\cite{YosinskiNIPS14How}.

Recent research has shown that explicitly reducing domain divergence upon the deep learning framework can further exploit domain invariant features~\cite{KrizhevskyNIPS12Imagenet}. A few big families have been identified in domain adaptation research~\cite{csurka17,wang18survey}. One family includes methods trying to match statistic moments on domain distributions, i.e. maximum mean discrepancy (MMD)~\cite{GrettonJMLR12Kernel} and its joint and conditional variants~\cite{LongTKDE16Deep,long2017cond}. 
Another family tries 
to align representations in source and target domains using optimal transportation~\cite{courty2017} or by associating source and target data in an embedding space~\cite{assocDA2017}.

The third family of methods is based on adversarial learning~\cite{cao2018-cvpr,cao2018-eccv,chadha2018,GaninJMLR16Domainadversarial,tzeng2017adda}.
Following the GAN principle~\cite{GoodfellowNIPS14Generative}, these methods encourage samples from different domains to be non-discriminative with respect to domain labels.

Ganin et al.~\cite{GaninJMLR16Domainadversarial} 
use a single domain classifier to regularize the extracted features to be indiscriminate with respect to the different domains. They assumed the existence of a shared feature space between domains where the distribution divergence is small. The domain-adversarial neural network (DANN) is integrated into the standard deep architecture to ensure that the feature distributions over the two domains are made similar. 
%
Adversarial discriminative domain adaptation (ADDA)
~\cite{tzeng2017adda} considers independent source and target feature extractors, 
where the parameters of the target model are initialized by the pre-trained source one. 
Other variations introduced different feature extractors for respective domains to learn more domain specific features~\cite{chadha2018}.

{\bf Class imbalance.}
\label{ssec:bias}
%
In the moment-based family of domain adaptation methods, Yan et al.~\cite{yan17} were first to explicitly take into account the class imbalance. They proposed a weighted MMD model that introduces an auxiliary weight for each class in the source domain.
The class weights are either known or estimated from iterative soft labeling of the target instances using the EM method. 






In the adversarial learning family, the open set adaptation adaptation labels allow new target classes and treat them as "unknown"~\cite{Saito_2018_ECCV},  partial domain adaptation 
relaxes the fully shared label space assumption to that the source label space subsumes the target label space. 
\cite{cao2018-cvpr} proposed the Selective Adversarial Network (SAN)
in order to distinguish between the relevant and irrelevant source classes. It tries simultaneously 
to down-weight the contribution of irrelevant classes when training the source classifier and domain discriminator, and to 
match the feature distributions in the shared label space.
To identify the importance score of source samples,~\cite{zhang2018cvpr} deploy two domain classifiers, where
the activations of first domain classifier are used as an indicator of the importance of each source sample to the target domain. By using these activations as weights, the weighted source samples and target samples are fed into the second domain classifier for optimizing the feature extractor.

The relevant problem of {\it selective bias} has been systematically studied in the semi-supervised learning~\cite{zadrozny_learning_2004} and deep learning with noisy labels~\cite{ren_learning_2018}. 
Selective bias refers to the difference between the training and test data. Most predictive models are sensitive to selective bias and current approaches try to correct this bias by adding pseudo-labeled data to the training sample, while instance weighting is used to make training sample similar to the distribution observed in the test data. 

Formulated as latent variable loss minimization, it can be solved by alternatively generating pseudo labels on target data and re-training the model with these labels~\cite{zou2018eccv}.
%
However, jointly learning the model and optimizing pseudo-labels on unlabeled data is naturally difficult as it is impossible to guarantee the correctness of the generated pseudo-labels. Semi-supervised learning can correct modest selective bias, but if the domain gap is too wide, initial predictions in the target domain will be poor,
thus increasing bias during training rather than decreasing it~\cite{frenay14}.
\section{Adversarial Domain Adaptation} 
\label{sec:lada}




The previous section identified two main approaches coping with the class imbalance in domain adaptation. The first approach is based on the source classifier 
to correctly predict the soft target labels~\cite{cao2018-cvpr,yan17,zou2018eccv}. 
However, if the gap between the domains is too large, iterative optimization with soft labels suffer from estimation error, bad initialization and convergence to local minima~\cite{frenay14}.

The second approach is to count on the discriminator and its activations as an indicator for re-weighting the source instances~\cite{zhang2018cvpr}.
However, under the class imbalance, {\it it is hard to distinguish between a poor domain discriminator and a low class probability in the target domain.}

We separate these two roles for the discriminator. First, the domain discriminator remains in charge of learning the separation between the source and target domains. Second, a new, auxiliary network is added to disentangle the target domain during learning, by injecting the latent codes and re-constructing them in semi-supervised GAN manner. 


\vspace{-1mm}
\subsection{Unsupervised Domain Adaptation}
\label{ssec:defs}
In unsupervised domain adaptation, source instances $X_s$ and labels $Y_s$ are drawn from a source domain distribution $p_s(x, y)$, while target instances $X_t$ are drawn from a target distribution $p_t(x)$, with no labels available. The goal is to learn such a target representation and classifier that can correctly classify target data into one of $K$ categories. 

Due to the domain shift, we have $p_s(x) \not= p_t(x)$.
Moreover, the source and target label spaces, ${\cal Y}^s$ and
${\cal Y}^t$, may differ.
In the {\it class alignment} case, the two spaces are identical, ${\cal Y}^t={\cal Y}^s$, and their class distributions are similar, $p^t(y) \approx p^s(y)$. In the {\it partial domain adaptation}~\cite{cao2018-cvpr,cao2018-eccv,zhang2018cvpr}, ${\cal Y}^t \subset {\cal Y}^s$, with $p_t(y)=0$ for classes which are in ${\cal Y}^s$ but not in ${\cal Y}^t$. In this paper, we consider the general case 
when ${\cal Y}^s$ is extended with empty "unknown" class for the {\it open set} case, and the presence of any source class in the target domain is not known. 
We pay attention to class imbalance when the target class distribution is very different from the source one, $p_t(y) \not= p_s(y)$. And we make no difference between the under-represented and absent classes, and use the KL divergence to measure the difference between the two distributions.

\vspace{-1mm}
\subsection{Adversarial discriminative domain adaptation}
\label{ssec:adda}   
We build on the previous adversarial learning works that combine a domain classifier with the domain representation learning to form the adversarial domain adaptation networks~\cite{cao2018-cvpr,chadha2018,chen_re-weighted2018,tzeng2017adda}. The goal is to learn both class discriminative and domain invariant representations, where the loss of the source classifier is minimized while the loss of the domain classifier is maximized. The minimax loss is similar to the original GAN:
\begin{equation}
\begin{tabular}{ll}
$\min\limits_{E_s,E_t} \max\limits_{D}{\cal L}=$ & $\E_{x\sim p_s(x)} [\log D(E_s(x))] +$ \\
&$\E_{x\sim p_t(x)} [\log(1-D(E_t(x)))]$,
\end{tabular}
\label{eq:gan}
\end{equation}
where $E_s$, $E_t$ are source and target encoders, $D$ is the domain classifier. $D$ corresponds to the discriminator in original GAN, with  the source and target instances labelled as 1 and 0, respectively. Maximizing (\ref{eq:gan}) with respect to $D$ yields a tighter lower bound on the true domain distribution divergence, while minimizing (\ref{eq:gan}) with respect to $E_s, E_t$ reduces the distribution divergence in the feature space. 

Since no labels are available in the target domain, the network is often trained in two stages~\cite{tzeng2017adda}. It first learns the source encoding $E_s$ and classifier $C$. Then it learns the adaptation to the target domain by regularizing the source and target encoders, $E_s$ and $E_t$, such that the distance between the source and target encoded representations $E_s(x)$ and $E_t(x)$ is minimized.

{\bf Source classifier.}
Source and target encoders are designed to capture domain specific representations.
In the first stage, the source classifier $C(E_s(x))$ 
is trained with the supervised loss and labeled source domain samples: 
\begin{equation}
\min\limits_{E_s,C} {\cal L}_{cls}= -\E_{x,y \sim p_s(x,y)}\sum_{k=1}^{K+1} {\bf 1}_{k=y} \log C(E_s(x)).
\label{eq:loss_superv}
\end{equation}

{\bf Adversarial losses.}
\label{ssec:adv_loss}
Once the source encoder $E_s$ is learned, a domain adversarial loss is used to reduce the discrepancy between two domains by optimizing the target encoder $E_t$ and discriminator $D$.
We follow~\cite{chadha2018,chen_re-weighted2018} in training the adversarial discriminator using the domain classification loss $L_{adv}^D$ and the encoder loss $L_{adv}^E$:
\begin{equation}
\begin{tabular}{ll}
$\min\limits_{D} {\cal L}_{adv}^D=$& $-\E_{x\sim p_s(x)} [\log D(E_s(x))] -$ \\
& $\E_{x\sim p_t(x)} [\log(1-D(E_t(x)))]$ \\
$\min\limits_{E_t} {\cal L}_{adv}^E=$& $-\E_{x_t \sim p_t(x)} [\log D(E_t(x_t))]$.
\end{tabular}
\label{eq:advD}
\end{equation}
The source encoder $E_s$ trained at the first stage is used to initialize the target encoder $E_t$. 

The discriminative loss above works well when classes in domains are aligned, $p_s(y) \approx p_t(y)$. In the case of class imbalance, the direct sampling $x\sim p_s(x), x\sim p_t(x)$ in (\ref{eq:advD}) is replaced with an instance re-weighting in the weighted ADDA~\cite{chen_re-weighted2018}, that estimates the $p(x_s)/p(x_t)$ ratio, or by detecting the outlier classes in partial domain adaptation, 
and down-weighting their contribution to the total loss~\cite{cao2018-eccv}. 
In the following section, we develop an alternative solution based on introducing the latent codes in the adversarial networks.

\subsection{Conditional GANs and latent codes}
\label{ssec:cgan}

Initially designed as generative models, 
GANs have been later extended to a conditional setting,
where a latent code $c$ is fed into the 
generator $G$ as an additional input layer~\cite{mirza_conditional_2014}.
The input noise $z$ and code $c$ are combined in a joint hidden representation, and the adversarial training framework allows for considerable flexibility in how this hidden representation is composed. 
InfoGAN~\cite{chen_infogan:_2016} is an extension that learns to maximize the mutual information between latent codes and observations. The latent code $c$ addresses the salient semantic features of the target data distribution.
To prevent generator $G$ from ignoring the latent codes $c$, InfoGAN regularizes learning via an additional cost term maximizing the mutual information $I(c,G(z,c))$ between the latent code $c$ and the generator output $G(z,c)$. 


{\bf Latent codes in adversarial domain adaptation}.
We apply latent codes to the adversarial domain adaptation framework 
presented in the previous section,
to disentangle the salient structure of target domain and to complement the source classifier $C$ and domain discriminator $D$ in estimating the target labels.

As the source encoder $E_s$ and classifier $C$ are not concerned with the latent codes, the classification loss ${\cal L}_{cls}$ in (\ref{eq:loss_superv}) and the first stage of training remain unchanged. 
At the second stage, 
the encoder representation $E(x)$, source or target, 
is now combined with a latent code $c$, as input $z=[E(x),c]$ to generator $G$. The code $c$ is then made meaningful by maximizing the mutual information $I(c;G(z))$ between the latent code $c$ and the generator output $G(z)$.
%
This mutual information can not be calculated explicitly, so that it 
is approximated using standard variational arguments~\cite{chen_infogan:_2016,spurr_guiding_2017}. 
This introduces an auxiliary classifier $Q$ 
modeled as a parameterized neural network, in order to approximate 
the likelihood of code $c$ given the input representation $E(x)$. 



To train the generator $G$, we sample the source and target instances
from their distributions $p_s(x)$ and $p_t(x)$, to form the input $z_s=[E_s(x),c]$, $z_t=[E_t(x),c]$ 
%
%
and use source labels $y$ as the latent codes, $c=y$. 
We additionally aim to increase the mutual information $I(c;G(z_t))$ between the latent codes and the unlabeled target samples. 

Training $Q$ network on labeled source data $(x,y)$ enables to encode the semantic meaning of labels $y$ via codes $c$ by increasing the mutual information $I(c;G(z_s))$. Simultaneously, the generator $G$ acquires the information of $y$ {\it indirectly} by increasing $I(c;G(z_t))$ and learns to utilize the  encoded representations of target instances. 

The updated form of the adversarial objective functions ${\cal L}_{adv}^D$ and ${\cal L}_{adv}^E$ in (\ref{eq:advD}) are given by combining the latent codes with the encoded representations of source and target samples: 
\begin{equation}
\begin{tabular}{ll}
$\min\limits_{D}{\cal L}_{adv}^{D}=$&$-\E_{x_s \sim p_s(x),c\sim {\cal P}_s} \log D(G([E_s(x_s),c]))$ \\
                           & $-\E_{x_t \sim p_t(x)_t,c\sim {\cal P}_t} \log (1-D(G([E_t(x_t)),c]))$\\
$\min\limits_{E_t}{\cal L}_{adv}^{E}=$&$-\E_{x_t \sim p_t(x), c\sim {\cal P}_t}\log D(G([E_t(x_t),c]))$
\end{tabular}
\label{eq:lada_d}
\end{equation}
  

The output of generator $G$ is fed to both domain discriminator $D$ and  
the auxiliary network $Q$; 
the later is trying to predict the class when feeding in the target input. The auxiliary minimization objective is obtained via Variational Information Maximization~\cite{chen_infogan:_2016} 
\begin{equation}
\min\limits_{Q} {\cal L}^{Q}= -\E_{x\sim p_d(x),c\sim {\cal P}_d} \log Q(c|G[E(x),c])+H(c),
\label{eq:lada_Q}
\end{equation}
where $d$ is domain indicator, $H(c)$ is the code entropy, $H(c) =\sum c_k \log_2 c_k$, $c_k$ is an estimate of class probability $y_k \in {\cal Y}$, $c_k=\frac{1}{N_t} \sum_i {\mathbb I}_{ik}$, $N_t$ is the number of target instances, ${\mathbb I}_{ik}$ is the indicator operator. Latent codes $c$ are sampled from a predefined and fixed distribution, ${\cal P}_s$ or ${\cal P}_t$, and therefore $H(c)$ is assumed to be a constant~\cite{chen_infogan:_2016}. 

\begin{figure}[ht]
\centering{\includegraphics[width=0.5\textwidth]{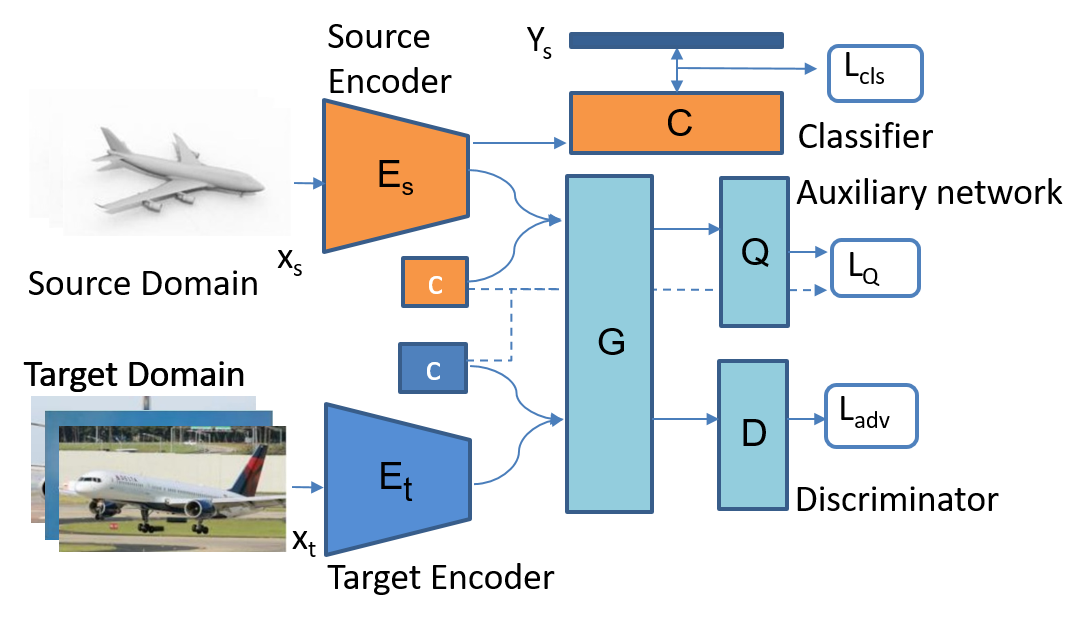}}
\caption{Latent code Adversarial Domain Adaptation (LADA) network.}
\label{fig:lada}
\end{figure}
The diagram of the proposed Latent code Adversarial Domain Adaptation (LADA for brevity) network is presented in Figure~\ref{fig:lada}. In addition to the source and target encoders $E_s, E_t$, classifier $C$ and domain discriminator $D$,  it includes latent codes $c$ 
and the auxiliary network $Q$ for the latent code reconstruction. 
In practice, $Q$ and $D$ share most convolutional layers and there is one final fully connected layer with the softmax function to output parameters for the conditional distribution $Q(c|z)$. 

{\bf Continuous relaxation of $H(c)$.}
\label{ssec:relax}
Under the class imbalance, we expect the distribution ${\cal P}_t$ for target domain to be estimated during the network training; this requires the entropy term $H(c)$ 
be differentiable in (\ref{eq:lada_Q}). 
%
However, $H(c)$ is discrete and non-differentiable. Below we derive a continuous relaxation of $H(c)$ and reformulate (\ref{eq:lada_Q}) as a fully variational minimization objective. 

$H(c)$ is non-differentiable due to the indicator operator ${\mathbb I}_{ik}$ which is 1 if target instance $x_i$ is labeled $y_k$ and 0 otherwise. We  smooth ${\mathbb I}_{ik}$ with a differentiable discrete probability distribution. We follow the entropy-constrained neural networks~\cite{wiedemann18} in introducing 
$P_{ik}=P(i=k|\theta_{ik})$ as the probability that the target instance $x_i$ takes the discrete value $y_k$, parametrized by $\theta_{ik}$. This continuous relaxation can model an entire set of target class distributions, whose probability of being sampled is specified by the joint probability distribution $P_\theta=\prod_{ik} P_{ik}$. Subsequently, we replace each class estimate $c_k=\frac{1}{N_t} \sum_i {\mathbb I}_{ik}$ with its continuous relaxation $P_k=\frac{1}{N_t} \sum_i P_{ik}$. The entropy term $H(c)$ is relaxed with $H(P)=\sum_k -P_k \log_2 P_k$, and the minimization objective~(\ref{eq:lada_Q}) is reformulated as 
\begin{equation}
\min\limits_{Q} {\cal L}^{Q}= -\E_{c\sim {\cal P}_d, x\sim p_d(x)} \log Q(c|G[E(x),c])+H(P).
\label{eq:relax}
\end{equation}

{\bf Three player game.} 
\label{par:3play}
So far, we followed the two stages training where source encoder $E_s$ and classifier $C$ are trained at the first stage, while the target encoder $E_t$ and discriminator $D$ are trained at the second stage. However, we also consider an alternative optimization, where all three components, $E_s$, $E_t$ and $D$, are trained jointly, by minimizing the LADA losses in (\ref{eq:lada_d})-(\ref{eq:lada_Q}).

We 
start with a moderately good source classifier $C$ and source encoder $E_s$, the later is used to initiate the target encoder $E_t$. At the second stage, we
alternate the minimization of three LADA losses: ${\cal L}^{cls}$, ${\cal L}_{adv}$ and ${\cal L}^{Q}$. This allows to back-propagate the gradient of the auxiliary loss ${\cal L}^{Q}$ to the source encoder $E_s$. The source encoder $E_s$ may evolve, sometimes paying a price of doing less well on source samples, towards a better domain invariant representation and latent code re-construction in $Q$ network. 

{\bf Alternative integration of latent codes.}
In LADA architecture, the choice was made to concatenate the latent codes with output of source and target encoders, $z=[E(x),c]$ when feeding to the discriminator $G$. 
There exist however other ways to inject the latent codes in the conditional GANs. These alternatives include a direct concatenation $z=[x,c]$ as input to both encoders, $E([x,c])$ or even using 
a dot-like operation~\cite{miyato18}. 

Our choice $z=[E(x),c]$ is motivated by the ease of integrating the latent codes in the discriminative adversarial network. No latent code is needed to classify a target instance with the source classifier as $C(E_t(x_t))$. Instead, in the $z=[x,c]$ case, the classification $C(E_t([x_t,c]))$ would require sampling the latent codes before encoding. 


\begin{algorithm}
\begin{algorithmic}[1]
\FOR {number of training iterations}
\STATE {Sample $m_t$ source and $m_t$ target pairs, $(x_s\sim p_s(x), c\sim {\cal P}_s)$, $(x_t \sim p_t(x), c\sim {\cal P}_t)$.}
\STATE {Update $D$ by ascending along its stochastic gradient: 
$\nabla_{\theta_D} [\frac{1}{m_s} \sum_{(x_s,c)}\log D(G([E_s(x_s),c])) +$ 
$\frac {1}{m_t} \sum_{(x_t,c)} \log (1- D(G([E_t[x_t,c]))]$}
\STATE Update $E_t$ by descending along its stochastic gradient: $\nabla_{\theta_{E_t}} \frac {1}{m_t} \sum_{(x_t,c)} \log D(G(E_t(x_t), c))$
\STATE Compute the relaxation  $H(P)$ of the entropy $H(c)$
\STATE Update $Q$ by descending along its stochastic gradient
$\nabla_{\theta_Q} [\frac{1}{m_s} \sum_{(x_s,c)} \log D(G[(E_s(x_s),c])) +$
$\frac{1}{m_t} \sum_{(x_t,c)} \log D(G[(E_t(x_t), c])) + {H}(P)]$
\ENDFOR
\end{algorithmic}
\caption{LADA stochastic gradient descent training.}
\label{alg1}
\end{algorithm}

\vspace{-0.5cm}
\section{Experiments}
\label{sec:eval}
The LADA network is implemented using TensorFlow Library 1.12 with CUDA 9.1. 
%
%
We train the LADA network on source images conditioned on their class labels and unlabeled target images. Both source and target encoders are fully connected layers. 
Domain encoded representations $E_s, E_t$ and latent code $c$ are mapped to generator $G$ composed of three hidden layers with the ReLu activation, with layer sizes 256, 128 and 64. The generator output is then mapped to 
fully connected layers for discriminator $D$ and auxiliary network $Q$, with 2 and $N_c$ variables, respectively, where $N_c$ is the number of source classes.

The model is trained using SGD (Agar optimizer) with mini-batches of size 32 and the initial learning rate of 0.001. 
Dropout with probability of 0.5 is applied to both domain encoders and the generator $G$. 
The classification accuracy on target images is used as the evaluation metric.

{\bf Extended evaluation protocol.}
\label{ssec:evaluation}
We extend the domain adaptation evaluation protocol beyond the cases of class alignment, 
open set or partial domain adaptation. 
The Kullback-Leibler divergence is a natural choice to measure the divergence between two 
distributions, and preferred to the class imbalance ratio~\cite{ortigosa2017}, a popular measure but applicable to one dataset only.
We therefore define the class imbalance ratio $CI$ as $KL(y_t||y_s)$. 
The KL divergence is defined only if for any class $c$, $p_s(y=c)=0$ implies $p_t(y=c)=0$. If $p_t(y=c)$ is zero, the contribution of the $c$-th term is interpreted as zero because $lim_{x \rightarrow 0^+} x \log (x)=0$. Therefore, it can cope with the both under-represented and absent target classes, when $p_t(y=c)$ is very low or simply zero




If the source and target classes are well aligned, the $CI$ values are close to zero.
In the partial domain adaptation evaluation on the Office 31 collection, 10 target classes are retained out of 31 available classes~\cite{cao2018-eccv}. For the six domain adaptation tasks, $CI$ values vary around 0.6. 



We adopt {\it random CI sampling} protocol which is an extension of two-class selective bias protocol~\cite{jiang2018mentonet} to the multi-class setting. We run a KL divergence generator and sample target subsets in such a way that $CI$ values 
vary between 0 and 1, we thus get a better idea of how UDA methods resist to the class imbalance.

\vspace{-2mm}
\begin{figure}[ht]
\begin{center}
\includegraphics[width=0.5\textwidth]{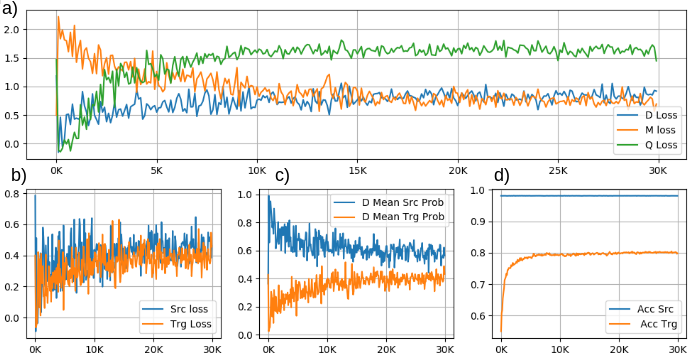}
\caption{LADA training: a) three LADA losses; b) Domain losses; c) Discriminator $D$ on target and source instances; c) Classification accuracy on source and target domains.} 
\label{fig:trace}
\end{center}
\end{figure}
\vspace{-2mm}
{\bf Evaluation options.}
\label{ssec:options}
We evaluate the performance of the LADA network presented in Section~\ref{sec:lada} under the class imbalance. 
The baseline is to disregard the latent codes in LADA network. By setting all latent codes to 0 in~Figure~\ref{fig:lada}, we down-scale the network to the plain ADDA net and denote it {\bf LADA-0}.



One option concerns the distributions ${\cal P}_s$, ${\cal P}_t$ for sampling the latent codes $c$.
Due to the semi-supervised latent codes in the network $Q$, we use the {\it coupled sampling} for the source encoder $E_s$.
When sampling source instances as input to generator $G$, 
$x\sim p_s(x),c\sim{\cal P}_s$, the encoded representation $E_s(x)$ 
is coupled with the latent code $c$ which is the corresponding label $y_s$\footnote{Note that $Q$ does not duplicate the source classifier $C$; it counts on the adversarial network to uncover the silent target structure.}. 
%
Concerning the distribution ${\cal P}_t$ when sampling latent codes $c$ for target samples (see Algorithm 1), we distinguish among three following options:
\begin{description}
\item[LADA-U] applies the uniform discrete distribution, ${\cal P}_t \sim Unif(K+1)$. 
\item[LADA-S] samples latent code $c$ from the source class distribution, 
${\cal P}_t = {\cal P}_s$. 
\item[LADA-T] combines the classifier $C$ and auxiliary network $Q$ to estimate the target class distribution ${\cal P}_t ={\hat p}_t(y)$. We first train the network with the uniform discrete distribution, $c~\sim Unif(K+1)$; then we switch to the the continual relaxation (\ref{eq:relax}),
continuously re-estimate the target class distribution $\hat p_t(y)$ and sample latent codes from it, $c\sim {\hat p}_t(y)$.
\end{description}

Another training option is on the second stage of LADA training. Along with the standard two stage training (LADA-2), 
we also consider the three player mode (LADA-3) which allows to back-propagate the auxiliary loss to the source encoder
and can learn better domain invariant representations.
 
Finally, to estimate the target class distribution ${\cal P}_t$, class probabilities for target instances, $p^C(y|x_t)$, predicted by classifier $C$ are complemented with $p^Q(y|x_t)$ by the auxiliary classifier $Q$, 
and used as the latent codes. We then simply average the two predictions. The product rule, commonly use in the class imbalance case, $p(y|x)=p^C(y|x) p^Q(y|x)/p(y)$, does not work as favoring the rare and absent classes. 

\vspace{-2mm}
\subsection{Datasets}
\label{ssec:data}
We run experiments on two image benchmark datasets. {\bf Office31} is a standard dataset for comparing visual domain adaptation. 
All domains in the collection are well aligned, with $CI$ values 
for the six source-target pairs lying between 0.03 and 0.08. 


%
{\bf VisDA} collection has been developed for the Visual Domain Adaptation Challenge 2017, with the main focus on the simulation-to-reality shift~\cite{visda2017}. In the image classification task, the goal is to first train a model on simulated, synthetic data in the source domain and then adapt it to perform well on real image data in the unlabeled test domain. The VisDA dataset is one of the largest for cross-domain classification, with over 280K source and 55K target images across 12 categories. 
We fine-tune ResNet-101 model pre-trained on ImageNet and use 2048 deep convolutional activation features as image representations.


 
 



\vspace{-2mm}

\begin{table*}[ht]
\begin{center}
\scalebox{0.75}{
\begin{tabular}{|l|ccc|ccc|ccc|ccc|ccc|ccc|ccc|ccc|ccc|ccc|ccc|} \hline
 Task &\multicolumn{3}{|c|}{$W \rar A$}
         &\multicolumn{3}{c|}{$W \rar D$}
              &\multicolumn{3}{c|}{$A \rar W$} 
                   &\multicolumn{3}{|c|}{$A\rar D$}
                      &\multicolumn{3}{c|}{$D\rar W$}
                          &\multicolumn{3}{c|}{$D\rar A$} &\multicolumn{3}{|c|}{Average} \\ \hline
$KL(S,T)$ &0.03 & 0.59& 1.0 & 0.08& 0.62 & 1.0&  0.04& 0.64 &1.0 & 0.07& 0.64&  1.0 & 0.08 & 0.63& 1.0   & 0.07& 0.64 &1.0   & 0.06& 0.63& 1.0 \\ \hline
OT & 77.5&56.6&45.9&   96.1&94.3&61.6&    84.5& 67.7& 60.2& 97.0&84.5& 75.3    &98.1& 96.3& 65.2& 92.0& \bf{91.2}& 62.2& 90.9& 82.6& 61.7  \\ 
ADDA10& 78.5&76.0&69.5&  98.2&96.3&68.2&    87.1& 72.2& 60.9&\bf{97.1}& 74.6& 65.1& \bf{98.2}& 97.1& 61.1&\bf{92.1}& 85.9& 60.2&\bf{91.6}& 81.5& 61.3  \\ 
SAN10    &-&\bf{83.2}&71.2&  -&\bf{100.0}&82.2& -&80.0& 72.9& -& 81.3& 71.3&  -  & 98.6& 75.1&  -  & 80.6& 72.5&  -  & 87.3& 74.2  \\ 
TDC10   &-& 81.7& 70.9&     -&\bf{100.0}&83.0& -&76.2& 73.1 & -& 79.0& 72.2&  -  &\bf{99.0}& 76.6&  -  & 89.5&\bf{73.4}&  -  & 87.6& 74.9  \\ 
LADA-30  &\bf{79.9}&78.2&70.2&\bf{98.6}& 98.3& 85.4& 87.1& 81.2 &74.2 & 95.3& 86.2& 73.9& 97.6& 91.0& 72.3& 88.6& 86.1& 66.2& 91.2& 86.8& 73.7 \\ 
LADA-3U   &79.7&78.1&71.5&    98.5&98.0& 89.2&  \bf{87.2}& 84.2 & 80.5& 95.4& 89.6& 77.2& 97.6& 95.3& 81.2& 88.6& 87.6& 71.3& 91.1& 88.8& 78.5 \\ 
LADA-3T  &79.7&77.9&\bf{72.0}&98.5& 98.1&\bf{90.1}& 87.1&\bf{85.5}&\bf{81.1} & 95.3&\bf{90.1}&\bf{79.0}& 97.8& 95.4&\bf{81.9}& 88.6& 87.6& 73.2& 91.2&\bf{89.1}&\bf{79.5} \\ \hline 
\end{tabular}
}
\caption{Results for OFF31 dataset: six (source,target) pairs and the average.}
\end{center}
\label{tab:off31}
\end{table*}

\subsection{Evaluation results}
\label{ssec:eval-e}
Figure~\ref{fig:trace} shows the LADA training process. Figure~\ref{fig:trace}.a tracks three LADA losses, ${\cal L}_{adv}^D$ (\textcolor{blue}{blue}), ${\cal L}_{adv}^{E}$ (\textcolor{myOrange}{orange}) and ${\cal L}_Q$ (\textcolor{green}{green}) over 30K iterations. Figure~\ref{fig:trace}.c shows how discriminator $D$ is good at distinguishing between the source and target domains. The initial ease of separating source domain (\textcolor{blue}{blue}) from the target one (\textcolor{orange}{orange}) vanishes through the iterations. The domain representations become more domain invariant, and $D$ faces more difficulty to distinguish between the domains, with the mean probability tending to 0.5 for either domain. Figure~\ref{fig:trace}.d tracks the accuracy of classifier $C$ on the source and target images. Due to the three player game, the classifier $C$ abandon a negligible fraction of accuracy on the source domain (\textcolor{blue}{blue}), in favor of a better target domain invariant representation. 

\begin{figure}[ht]
\begin{center}
\includegraphics[width=0.5\textwidth]{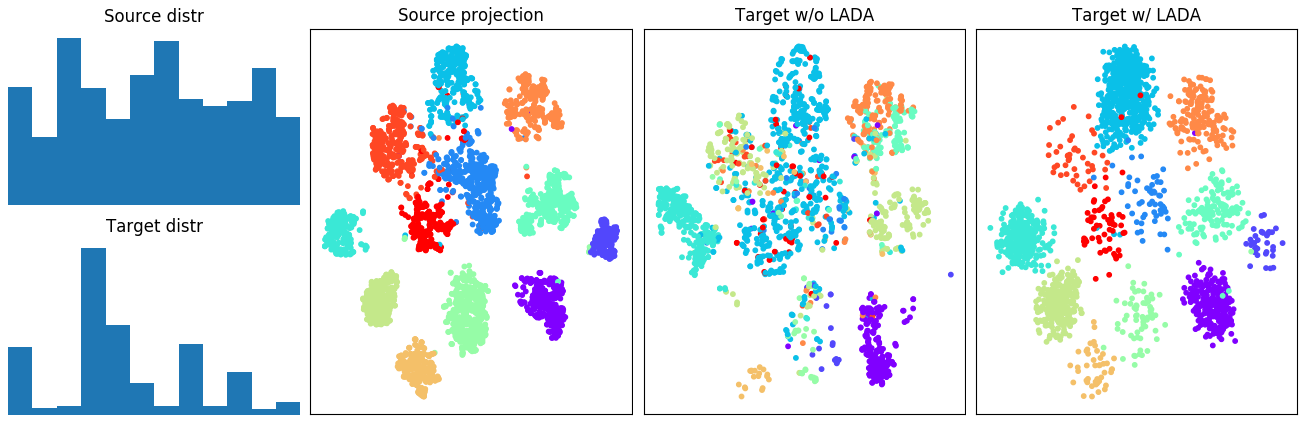} 
\caption{Severe class imbalance in VisDA set. T-SNE projections without and with LADA net.} 
\label{fig:tsne}
\end{center}
\end{figure}
\vspace{-2mm}
\subsection{Qualitative and quantitative results}
\label{ssec:quality}
We revisit the T-SNE projection example in Figure~\ref{fig:ci2} where the baseline UDA method was unable to separate small and big classes under the severe class imbalance ($CI$=0.71). As Figure~\ref{fig:tsne} shows, using LADA network helps accurate estimate the target class distribution and separate small classes from big ones.

\vspace{-2mm}
\paragraph{Office 31.}
We run experiments with three LADA versions for three different $CI$ values, given by the dataset split ($CI$ in [0.03:0.08]), the partial class alignment ($CI$ in [0.59:0.63]) and the severe class imbalance ($CI$=1.0). Classification accuracy for the six domain adaptation tasks are presented in Table~4. 
For each task, we average over 10 runs 
when training LADA-30, LADA-3U and LADA-3T models. The table also averages the accuracy values for all six domain adaptation tasks. 

All three LADA versions perform similarly when $CI$ values are inferior to 0.6. Their performance is comparable to the state of art methods. In the class alignment case, they are compared to ADDA~\cite{tzeng2017adda} and Optimal Transport~\cite{courty2017}.
In partial domain adaption case, they are compared to the weighted ADDA (ADDA10)~\cite{chen_re-weighted2018}, Selective Adversarial networks (SAN10)~\cite{cao2018-cvpr} and Two Discriminator Network (TDN10)~\cite{zhang2018cvpr}.
When the CI values are close to 0.6, LADA-3U and LADA-3T behave on average better than methods specifically developed for the partial domain adaptation. The LADA methods however make no assumption on the absence of some classes; they treat equally cases when the classes are under-represented or absent. 
For all methods, performance starts decreasing when the $CI$ values approach 1. 
LADA-3U and LADA-3T resist better to the class imbalance than all other methods.
\begin{table}[ht]
\begin{center}
\scalebox{0.75}{
\begin{tabular}{|l|rrrrr|} \hline
 Task &\multicolumn{5}{|c|}{$VisDA$} \\ \hline
$KL(S,T)$&0.05 &0.3 &0.6 &0.8 &1.0 \\ \hline
LADA-0  &84.9&     84.9&     83.7&   77.1  & 63.1\\
LADA-2U &85.1&     84.5&     84.3&   81.7  & 78.6 \\
LADA-3S &\bf{85.2}&\bf{85.4}&84.0&   82.7  & 79.7 \\
LADA-3U &84.5&     84.7&     83.9&   82.9  & 80.5 \\
LADA-3T &85.1&     84.9& \bf{84.5} &\bf{83.6} &\bf{81.6} \\ \hline
\end{tabular}
}
\caption{Results for VisDA dataset.}
\end{center}
\label{tab:visda}
\end{table}

\vspace{-2mm}
\paragraph{VisDA.} 
We use this dataset to run ablation study and investigate different options of the LADA training; the evaluation results are reported in Table~\ref{tab:visda}. We compare LADA-3 to LADA-2 and the baseline. As the figure shows, the LADA-0 does not resist to the severe class imbalance. The difference between LADA-2 and LADA-3 is small but statistically significant. The three player game allows to maintain the target accuracy, for the price of performing slightly worse on the source samples. By optimizing the source encoder $E_s$ at the second stage, the classifier $C$ can benefit from the disentangling the target domain with the auxiliary classifier $Q$.

{\bf Class imbalance reduction.} 
\label{ssec:kl-track}
In the last ablation experiment, we study how good the LADA models are at estimating the (unknown) target class distribution.
Given an initial $CI$ value, we train LADA-3U and measure how the target class estimation ${\hat p}_t(y)$ diverges from the true target class distribution $p_t(y)$. Again, we use the KL divergence to measure it, $KL({\hat p}_t(y)||p_t(y))$ (\textcolor{red}{red}). We also track the divergence of ${\hat p}_t(y)$ from the source distribution $p_s(y)$ (\textcolor{green}{green}).

\begin{figure}[ht]
\begin{center}
\includegraphics[width=0.45\textwidth]{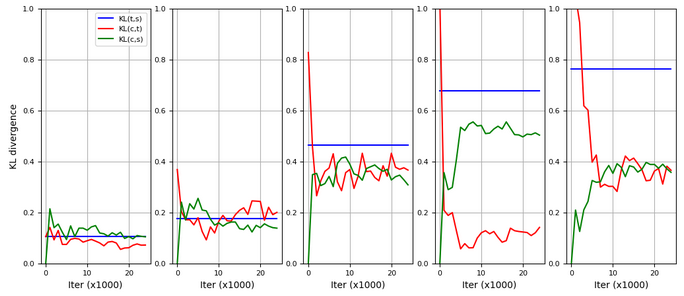}
\caption{KL-divergence between the true and estimated target class distributions, for different $CI$ values.}
\label{fig:kl-conv}
\end{center}
\end{figure}

Figure~\ref{fig:kl-conv} reports LADA-3U performance on the class imbalance reduction, with the starting $CI$ values that grow from 0.1 to 0.9 (\textcolor{blue}{blue}). It is able to reduce the divergence between ${\hat p}_t(y)$ and $p_t(y)$ in most cases. However, there exists an incompressible divergence, that none model seems to be able to break up. 

\vspace{-5mm}
\section{Conclusion}
\label{sec:conclusion}
We propose a novel adversarial network for unsupervised domain adaptation, specific for severe class imbalance when the class distributions in the source and target domains diverge considerably. We address real world scenarios when source classes are under-represented of absent at all in the target domain. We test domain adaptation methods under the class imbalance using  an extended evaluation protocol. We use latent codes in the adversarial learning aimed at disentangling the salient structure of the target domain. The auxiliary network is introduced in the adversarial domain adaptation framework to reconstruct latent codes; it plays the role of additional predictor when facing the severe class imbalance.


\begin{footnotesize}
\bibliographystyle{aaai}
\bibliography{DomAd-all,mybib-da1,MyLibrary,DomAd}
\end{footnotesize}
\end{document}